\documentclass[10pt,twocolumn,letterpaper]{article}

\usepackage[pagenumbers]{wacv} 

\usepackage{graphicx}
\usepackage{amsmath}
\usepackage{amssymb}
\usepackage{booktabs}
\usepackage{tabularx}
\usepackage[table,xcdraw]{xcolor}
\usepackage{adjustbox}

\usepackage[pagebackref,breaklinks,colorlinks]{hyperref}

\usepackage[capitalize]{cleveref}
\crefname{section}{Sec.}{Secs.}
\Crefname{section}{Section}{Sections}
\Crefname{table}{Table}{Tables}
\crefname{table}{Tab.}{Tabs.}

\def\wacvPaperID{} 
\def\confName{WACV}
\def\confYear{2025}

\DeclareMathOperator*{\bigconcat}{Concat}
\begin{document}

\title{GMT: Guided Mask Transformer for Leaf Instance Segmentation}

\author{
Feng Chen\textsuperscript{1} \quad Sotirios A. Tsaftaris\textsuperscript{1} \quad Mario Valerio Giuffrida\textsuperscript{2} \\
\textsuperscript{1}IDCOM, School of Engineering, University of Edinburgh \\
\textsuperscript{2}School of Computer Science, University of Nottingham 
\\
{\tt\small feng.chen@ed.ac.uk, s.tsaftaris@ed.ac.uk, valerio.giuffrida@nottingham.ac.uk}
}
\maketitle

\newcommand{\mypar}[1]{\noindent\textbf{#1.}}

\begin{abstract}
Leaf instance segmentation is a challenging multi-instance segmentation task, aiming to separate and delineate each leaf in an image of a plant. Accurate segmentation of each leaf is crucial for plant-related applications such as the fine-grained monitoring of plant growth and crop yield estimation. This task is challenging because of the high similarity (in shape and colour), great size variation, and heavy occlusions among leaf instances. Furthermore, the typically small size of annotated leaf datasets makes it more difficult to learn the distinctive features needed for precise segmentation. We hypothesise that the key to overcoming the these challenges lies in the specific spatial patterns of leaf distribution. In this paper, we propose the Guided Mask Transformer (GMT), which leverages and integrates leaf spatial distribution priors into a Transformer-based segmentor. These spatial priors are embedded in a set of guide functions that map leaves at different positions into a more separable embedding space. Our GMT consistently outperforms the state-of-the-art on three public plant datasets. Our code is available at \url{https://github.com/vios-s/gmt-leaf-ins-seg}.
\end{abstract}

\section{Introduction}
\label{sec:intro}
Plant phenotyping measures the structural and functional traits of plants such as leaf area and flowering time, playing a key role in various agricultural domains including breeding and crop management \cite{pieruschka2019plant}. Quantifying leaf traits is crucial for understanding plant functions such as respiration, nutrition, and photosynthesis \cite{bhagat2022eff}. Recent advancements in computer vision and machine learning have enhanced plant analysis for phenotyping and accelerated scientific discoveries for the plant community \cite{LI2020105672}. This paper focuses on the image-based analysis of plant leaves as a non-destructive approach to plant phenotyping.
\begin{figure}[t]
     \centering
    \includegraphics[width=0.95\linewidth]{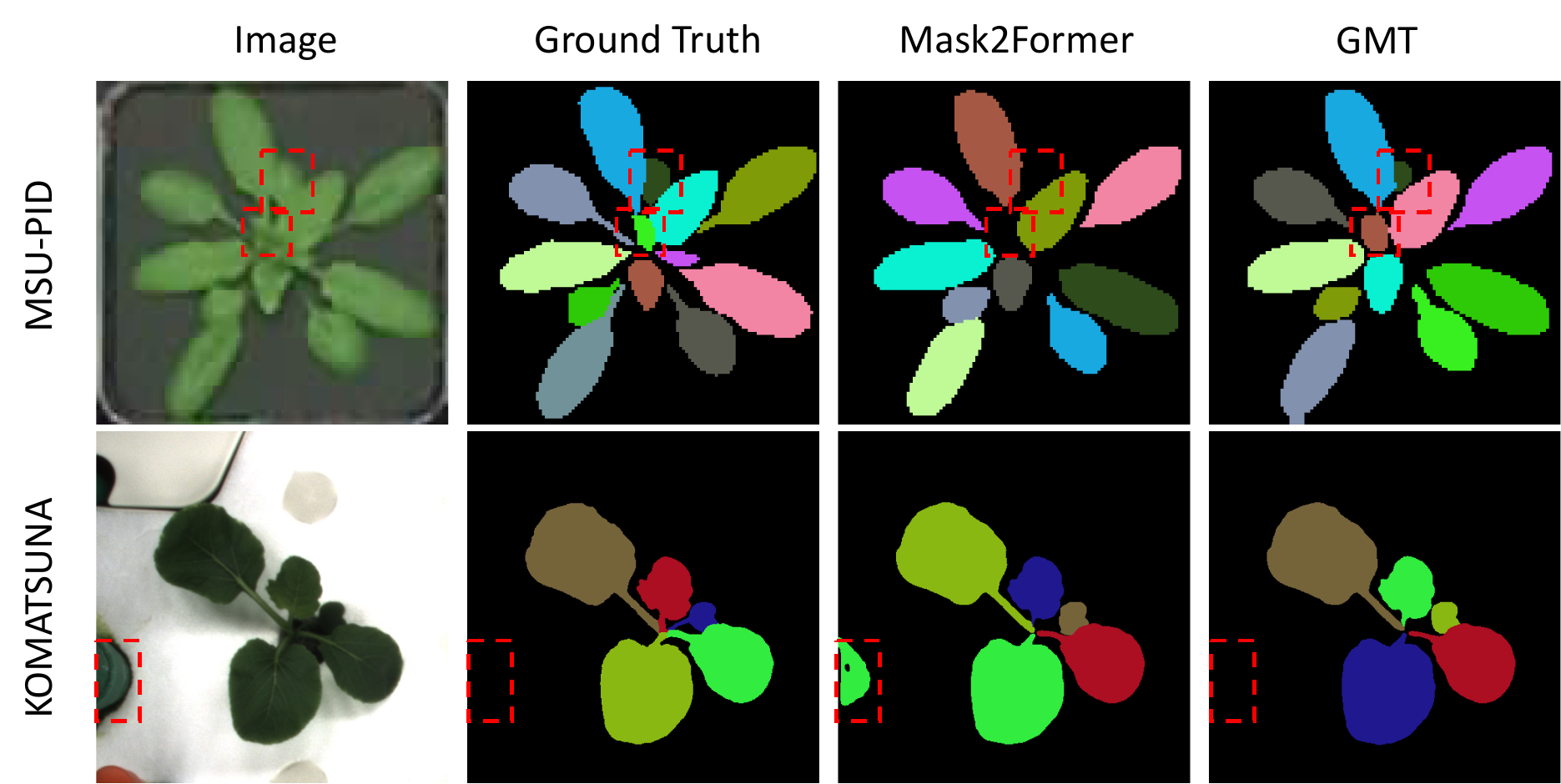}
    \caption{Examples of leaf instance segmentation. Each row presents an example from a different dataset, showing the original image, ground-truth label, Mask2Former \cite{cheng2022masked} prediction, and our proposed GMT segmentation result, from left to right.}
    \label{fig:ins_seg_example}
\end{figure}

Instance segmentation is a well-studied computer vision task aiming at precisely delineating each object within an image at the pixel level \cite{hafiz2020survey, minaee2021image}. While state-of-the-art (SOTA) Transformer models \cite{vaswani2017attention, dosovitskiy2020image, carion2020end} have outperformed convolutional neural networks (CNNs) in various tasks, they still struggle to achieve robust instance segmentation for plant leaves. This challenge is caused by the small size of most plant datasets, which typically contain only a few hundred to a few thousand images, making it difficult to train large models effectively. Moreover, plants exhibit severe overlaps and a huge intra- and inter-species morphological variability \cite{scharr2016leaf}, which further challenges computer vision approaches. As illustrated in \cref{fig:ins_seg_example}, current Transformer-based segmentors such as Mask2Former \cite{cheng2022masked} (column 3) often fail to segment leaves correctly, missing small and overlapping leaves (row 1) or confusing other green objects with leaves (row 2). The challenges arising from leaf instance segmentation have been extensively studied for years in the CVPPP/CVPPA series of workshops.\footnote{\url{https://cvppa2021.github.io/challenges/}}

In many plants, particularly rosette plants (\eg Arabidopsis), leaves grow outward from the centre, with those near the centre being younger (typically smaller and more densely clustered), while those farther out are older and tend to be larger \cite{giuffrida2016learning}. Also, leaves at different growth stages exhibit distinct shapes. These features can be captured via top-down views of plants (examples in \cref{fig:ins_seg_example}) using laboratory imaging devices or UAVs \cite{huther2020}.

In other words, leaf positions are correlated to the challenges in plant leaf instance segmentation. This correlation motivates us to incorporate leaf position as prior knowledge to improve segmentation models. Therefore, we present the Guided Mask Transformer (GMT), a Transformer-based model tailored to instance segmentation on plant images (\cref{fig:gmt}). Our model introduces three novel components, namely Guided Positional Encoding (GPE), Guided Dynamic Positional Queries (GDPQ), and Guided Embedding Fusion Module (GEFM), to integrate prior knowledge on the spatial distributions of leaf instances which is represented by guide functions. Specifically, GPE and GDPQ enhance the positional representation of instances within the attention mechanism, while GEFM improves the separability of feature representations for different instances.

Experimental results on three popular plant datasets \cite{minervini2016finely, cruz2016multi, uchiyama2017easy} demonstrate that GMT outperforms SOTA leaf instance segmentation models. Quantitative and qualitative results show that the performance improvements are most notable for small to medium-sized leaves, especially for the overlapping ones (\cref{fig:ins_seg_example} top right), and the confusion with leaf-like objects is reduced (\cref{fig:ins_seg_example} bottom right). This highlights our method's ability of addressing the most challenging scenarios in plant analysis. We also present ablation studies for GMT's key components, demonstrating that its overall design is crucial to the learning process.

In summary, our  key contributions are: (i) we introduce GMT, a Transformer-based segmentor that leverages position-related prior knowledge to improve leaf instance segmentation; (ii) we show SOTA performance on three public datasets and discuss its ability to tackle the most challenging scenarios in plant data analysis, with ablation studies validating the importance of its overall design.

\section{Related Work}

\mypar{Plant Image Segmentation}
Plant image segmentation is an important step in many agricultural and plant phenotyping applications \cite{jiang2020convolutional, zhang2020applications}. CNN-based models have been prominent, such as Singh \& Misra \cite{singh2017detection} using segmentation to detect leaf diseases, Zhang \etal \cite{zhang2022wheat} proposing a wheat spikelet segmentation method based on the hybrid task cascade model \cite{chen2019hybrid}, and Jia \etal \cite{jia2021foveamask} introducing FoveaMask for fast instance segmentation of green fruits. More recently, Transformer-based models have gained attention in plant image segmentation \cite{chen2023adapting, du2023pst, jiang2022transformer, chen2023pctrans}, though there is still a gap in effectively incorporating plant-specific knowledge for improved segmentation.

\mypar{Transformer-based Segmentation}
Transformers has been successfully applied across various domains of machine learning, including NLP \cite{devlin2018bert, brown2020language} and computer vision \cite{dosovitskiy2020image, bao2021beit, peebles2023scalable}. With the success of Detection Transformer (DETR) \cite{carion2020end} in object detection, subsequent studies have adopted similar principles to tackle image segmentation. For example, MaskFormer \cite{cheng2021per} leverages a Transformer decoder to refine a set of object queries via cross attention and self attention, and these refined object queries are used to produce semantic masks. Mask2Former \cite{cheng2022masked} proposes the masked cross attention to reduce computational burden and increase segmentation quality, and extends MaskFormer's ability to semantic, instance, and panoptic segmentation. Furthermore, He \etal \cite{he2023dynamic} introduce focus-aware dynamic positional queries alongside a method for performing cross-attention with high-resolution feature maps, both of which are crucial for image segmentation. Furthermore, FastInst \cite{he2023fastinst} introduces a framework for real-time instance segmentation, while OneFormer \cite{jain2023oneformer} is a model trained once to handle multiple segmentation tasks simultaneously.

Despite advancements in Transformer-based segmentation models in general domains, challenges remain in plant image analysis due to plant complexity and small datasets. To address this, we propose the Guided Mask Transformer (GMT), which integrates prior knowledge of leaf and plant characteristics to enhance leaf instance segmentation.

\section{Proposed Method}
\label{sec:methodology}
\begin{figure*}[t]
    \centering
    \includegraphics[width=0.9\linewidth]{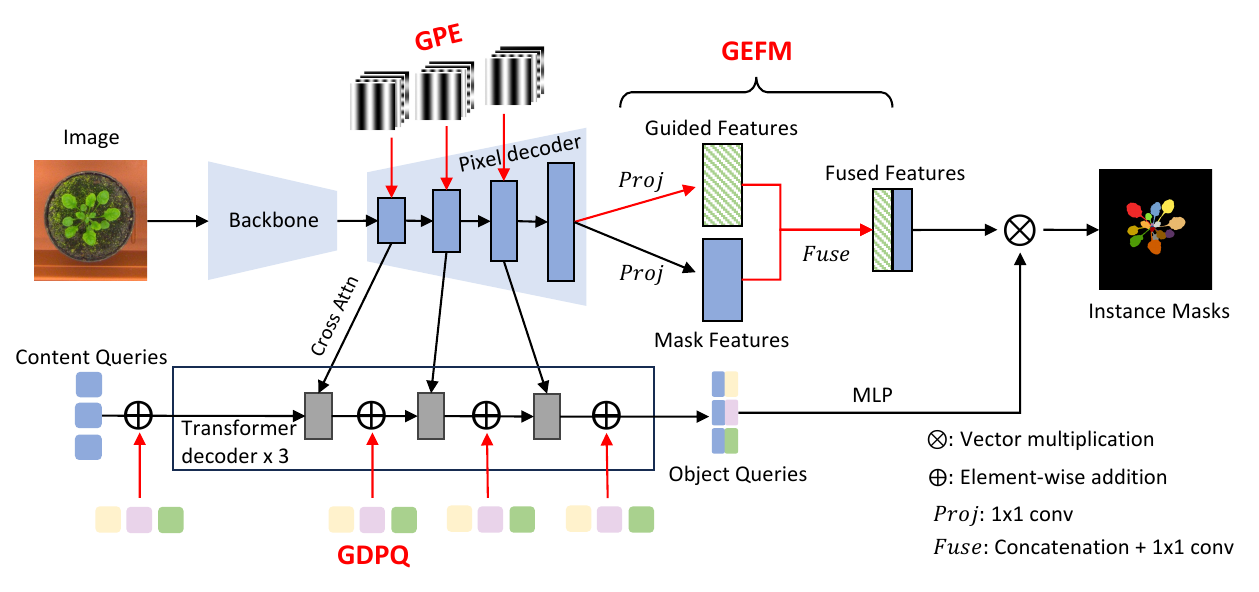}
    \caption{An illustration of the Guided Mask Transformer (GMT). The key components of GMT are Guided Positional Encoding (\textcolor{red}{\textbf{GPE}}), Guided Embedding Fusion Model (\textcolor{red}{\textbf{GEFM}}), and Guided Dynamic Positional Queries (\textcolor{red}{\textbf{GDPQ}}). These components enable an effective integration with the guide functions carrying prior knowledge on instances' distribution. (Best viewed in colour.)}
    \label{fig:gmt}
\end{figure*}

The Guided Mask Transformer (GMT) (\textit{c.f.} \cref{fig:gmt}) captures and represents unique leaf and plant characteristics, integrating them into a segmentation model. Using guide functions tailored to pixel coordinates, we optimise these functions to distinguish instances within an embedding space. Then, GMT  leverages these optimised guide functions with instance distribution priors to predict plant leaf instance masks.

\subsection{Guide Functions}
\label{sec:guide_functions}
We adopt the harmonic functions \cite{kulikov2020instance} as guide functions. These harmonic functions, which are adjusted to instance locations and able to separate distinct instances in an embedding space (after training), are defined as:
\begin{equation}
    \label{eq:guide}
    f_{i}(x, y; \psi_{i}) = \sin \left( \frac{\psi_{i}{\left[1\right]}}{W} x + \frac{\psi_{i}{\left[2\right]}}{H} y + \psi_{i}{\left[3\right]} \right),
\end{equation}
where $\psi_{i}{\left[1\right]}$ and $\psi_{i}{\left[2\right]}$ are the learnable frequency parameters, $\psi_{i}{\left[3\right]}$ is the learnable phase parameter; $W$ and $H$ represent the image size; $x$ and $y$ denote the pixel coordinate. Given a set of pixels $S$ belong to an instance, we can calculate the expectation of $f_{i}$ over this instance:
\begin{equation}
    \label{eq:ef}
    e_i(S; \psi_i) = \frac{1}{|S|} \sum_{(x, y) \in S} f_i(x, y; \psi_i).
\end{equation}

Assuming the number of guide functions is $d_{g}$, the embedding of an instance (we refer it as the guided embedding following \cite{kulikov2020instance}) is represented as a joint vector:
\begin{equation}
    \label{eq:ge}
    e(S; \Psi) = \left\{e_1(S; \psi_1), e_2(S; \psi_2), \ldots, e_{d_{g}}(S; \psi_{d_{g}}) \right\}.
\end{equation}

The guide functions are trained to separate the guided embedding of different instances to a pre-set distance $\epsilon$, and the loss function is defined as:
\begin{align}
    \label{eq:guide_loss}
    \ell(\Psi) &= \sum_{I \in \mathcal{I}} \frac{1}{|P_I|} \sum_{(S, S') \in P_I} \notag \\
    &\quad \max \left(0, \epsilon - \|e(S; \Psi) - e(S'; \Psi)\|_{1} \right),
\end{align}
where $\mathcal{I}$ denotes the set of training images, $P_I$ represents all possible pairs of instances within an image $I$, $S$ and $S'$ are pixels belong to two different instances, and $\|\cdot\|_{1}$ is the $L_1$ distance.

After training, we obtain $d_g$ guide functions, each with its own learned parameters $\psi_i$ (\textit{c.f.} \cref{eq:guide}). These guide functions are adjusted to the pixel coordinates of different instances (\textit{c.f.} \cref{eq:guide}), allowing them to carry prior information of instance locations; they are also able to embed distinct instance masks and separate them in the embedding space (\textit{c.f.} \cref{eq:guide_loss}).

\subsection{GMT: Guided Mask Transformer}
Once the guide functions are obtained, we incorporate them into the proposed GMT model. Our key contribution lies in the novel modules we introduce, GPE, GEFM, and GDPQ, to fully leverage these functions within a Transformer-based architecture, as detailed at \cref{subsubsec:gmt}. We extend the meta-architecture of Mask2Former \cite{cheng2022masked} with our proposed modules, given its strong instance segmentation capabilities in general domains but limitations in leaf instance segmentation (\textit{c.f.} \cref{fig:ins_seg_example} column 3). 

\subsubsection{Mask2Former Preliminaries}
\label{sec:mask2former}
Mask2Former consists of a backbone to extract image features, a pixel decoder to generate multi-scale pixel features, and a Transformer decoder to refine object queries through cross-attention (with the multi-scale pixel features) and self-attention. Instance masks are predicted by multiplying the refined object queries with the mask features projected from the highest-resolution pixel features. Our proposed modules integrating leaf instance spatial priors primarily operate within the pixel decoder and Transformer decoder.

The \textbf{pixel decoder} gradually upsamples features extracted by the backbone, producing detailed high-resolution features for each pixel. The default pixel decoder of Mask2Former is a multi-scale deformable attention Transformer (MSDeformAttn) \cite{zhu2020deformable}, with pixel features at scale $s$ denoted as $K^s = K^s_c + K^s_p$, where $K^s_c$ are the content features, and $K^s_p$ are the corresponding positional encodings.

The \textbf{Transformer decoder} progressively refines a set of object queries $Q$ through attention mechanisms. These queries are formed as the addition of content queries $Q_c$, which capture the semantic information of objects, and positional queries $Q_p$, which encode the spatial information of instances. A key aspect of this query refinement process is using cross-attention, where each query attends to all pixel features from the pixel decoder to refine its representation. Below we briefly introduce the cross-attention operation.

Let $n$ be the number of queries; $h$, $w$, and $d_p$ are the pixel features' height, width and dimension, respectively. We denote the pixel features (omitting the scale level) as $K \in \mathbb{R}^{hw \times d_p}$, and the object queries as $Q \in \mathbb{R}^{n \times d_p}$. The cross-attention operation is expressed as follows:
\begin{equation}
\label{eq:cross-attention}
\operatorname{Crs-Attn}(Q, K, V) = \operatorname{softmax}\left(\frac{QK^{T}}{\sqrt{d_p}}\right) V,
\end{equation}
where $V \in \mathbb{R}^{hw \times d_p}$ is identical to $K$, and the first term is known as the cross-attention map, \( A = \operatorname{softmax}\left(\frac{QK^{T}}{\sqrt{d_p}}\right) \). \cref{eq:cross-attention} shows that cross-attention aggregates image context based on the dot-product similarity between the queries $Q$ and keys $K$, Both the content and positional components of $Q$ and $K$ contribute to the attention scores, leveraging both semantic and spatial information.

\subsubsection{Guided Mask Transformer}
\label{subsubsec:gmt}
Our GMT follows the standard design of Mask2Former, using MSDeformAttn \cite{zhu2020deformable} as the pixel decoder and a Transformer decoder composed of nine blocks, each with a masked cross-attention layer, a self-attention layer, and a feed-forward layer. Intermediate supervision is applied after each Transformer block predicting instance masks. Below we describe the key components of GMT, which replace or enhance parts of the pixel decoder and Transformer decoder. These adaptations enable the effective integration of the guide functions obtained from the previous stage. The overall architecture of GMT is shown in \cref{fig:gmt}.

\mypar{Guided Positional Encoding (GPE)}
As introduced in \cref{sec:mask2former}, the positional components of pixel features in the pixel decoder enhance spatial representation of instances. We propose GPE to better align this spatial representation with the prior knowledge of leaf instance distribution.

By default, sinusoidal positional encoding (SPE) (as introduced in \cite{vaswani2017attention}) is added to the multi-scale pixel features at the pixel decoder in Mask2Former. SPE is defined as:
\begin{align}
    \label{eq:spe}
    \operatorname{SPE}_{pos, 2j} &= \sin\left(\frac{pos}{10000^{2j/d_{p}}}\right), \notag \\
    \operatorname{SPE}_{pos, 2j+1} &= \cos\left(\frac{pos}{10000^{2j/d_{p}}}\right),
\end{align}
where $d_{p}$ is the dimension of pixel features, $j\in \left[ 0, d_{p}/2-1\right]$ indexes the dimension of SPE and $pos$ denotes pixel coordinates (\ie $x$ or $y$). Typically, $pos=x$ when $j<d_{p}/4$, otherwise $pos=y$.

As presented in \cref{sec:guide_functions} and \cref{eq:guide}, the guide functions are sinusoidal, correspond to the pixel coordinates of instances, and possess positional priors of leaf instances after training. This makes them suitable for improving the spatial representation of the original SPE. However, the number of guide functions $d_{g}$ is much smaller than the pixel feature dimension $d_{p}$; for instance, in our experiments, $d_{g}=16$ and $d_{p}={256}$. An intuitive approach is to increase $d_g$ to $d_p$ when learning the guide functions. However, we found this to be impractical due to high computational cost and optimisation challenges for the guide functions at such high dimensionality. Therefore, we propose a ``postprocessing'' method to expand the low-dimensional guide functions to high-dimensional positional encodings. We expand the number of guide functions to $d_{p}$ by maintaining the frequency components $\psi{\left[1\right]}$ and $\psi{\left[2\right]}$, while shifting the phase components $\psi{\left[3\right]}$. Specifically, every guide function $f_{i}$ is expanded to $J = d_{p} / d_{g}$ functions, defined as:
\begin{align}
    f_{i,k}(x, y; \psi_{i}, j) = \sin \Bigg( & \frac{\psi_{i}{\left[1\right]}}{W} x + \frac{\psi_{i}{\left[2\right]}}{H} y \notag \\
    & + \psi_{i}{\left[3\right]} + 2\pi\frac{d_{g}}{d_{p}}j \Bigg),
\end{align}
where $j=0,1,2,\ldots,J-1$. The expanded set of guide functions is denoted as:
\begin{align}
    \mathcal{F} = \Big\{ & f_{0,0}, f_{0,1}, \ldots, f_{0,J-1}, f_{1,0}, f_{1,1}, \ldots, f_{1,J-1}, \ldots, \notag \\
    & f_{d_{g} - 1,0}, f_{d_{g} - 1,1}, \ldots, f_{d_{g} - 1,J-1} \Big\}.
\end{align}
The concatenation of all elements in $\mathcal{F}$ is then added to SPE to form GPE. GPE combines both absolute pixel coordinate information and pre-learned instance spatial distributions, and enhance spatial representation of instances.

\begin{figure}[t]
    \centering
    \includegraphics[width=1\linewidth]{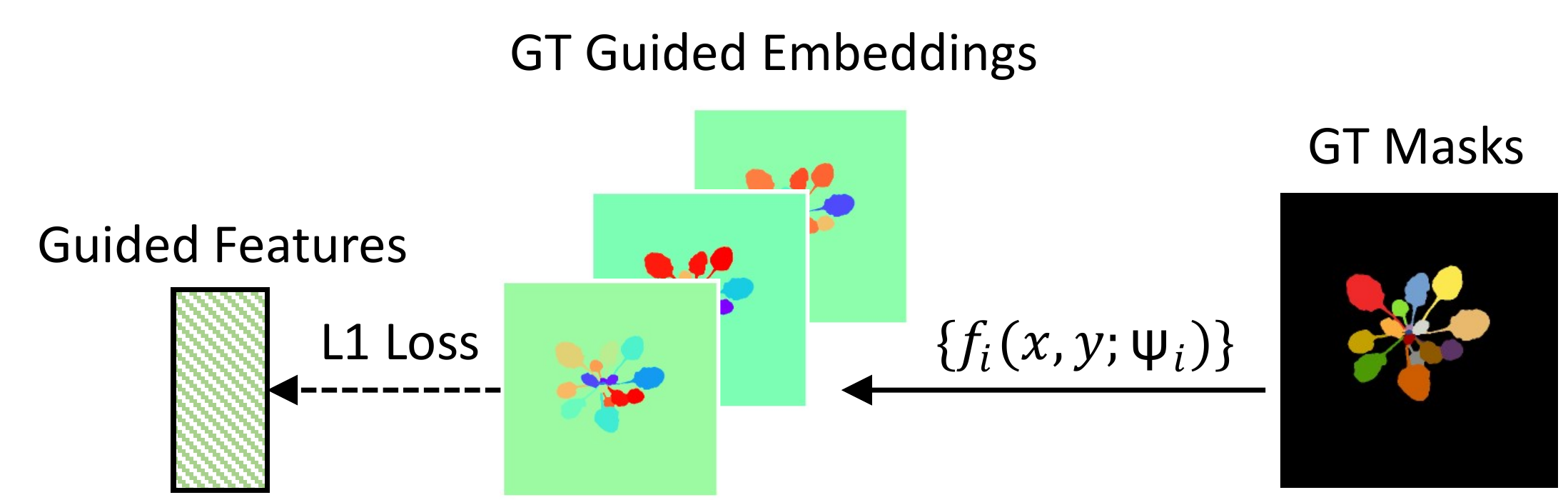}
    \caption{Auxiliary supervision at GEFM. The ground-truth (GT) instance masks are encoded by the trained guide functions to produce the GT guided embeddings. The guided features, which are projected from the final output of pixel decoder, are supervised by these embeddings with an $L_1$ loss.}
    \label{fig:auxsup}
\end{figure}

\begin{figure*}[t]
    \centering
    \includegraphics[width=0.85\linewidth]{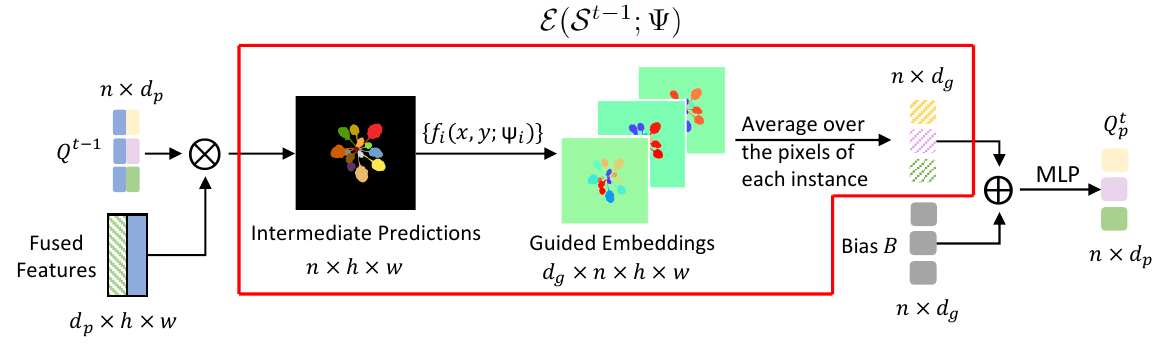}
    \caption{GDPQ Module. The positional queries at current Transformer block $Q^{t}_{p}$ is dynamically generated on the guide-function-encoded mask predictions from last block. $Q^{t-1}$ denotes the object queries from last Transformer block. The dimensions of different elements are shown: $n$ is the length of object queries, and $h$ and $w$ denote the spatial dimension of the final pixel features. The computation process of $\mathcal{E}(\mathcal{S}^{t-1}; \Psi)$ as presented in \cref{eq:gdpq} is highlighted in \textcolor{red}{red}.}
    \label{fig:gdpq}
\end{figure*}

\mypar{Guided Embedding Fusion Module (GEFM)}
As shown in \cref{eq:guide_loss}, the trained guide functions can map the instance masks to guided embeddings where different instances are well separated. To facilitate this mapping in GMT and enhance the distinction between similar leaf instances, we use a $1\times1$ convolution layer to project the pixel decoder outputs to the guided features (\cref{fig:gmt}). These guided features are supervised by the ground-truth guided embeddings (obtained by encoding the ground-truth instance masks with the trained guide functions), with an $L_1$ loss\footnote{The loss is weighted higher on the edges of instances, following the implementation of \cite{kulikov2020instance}.}, as shown in \cref{fig:auxsup}. Another $1 \times 1$ convolution layer is used to fuse the concatenation of guided features and the original mask features (\cref{fig:gmt}).

\mypar{Guided Dynamic Positional Queries (GDPQ)}
As discussed in \cref{sec:mask2former}, the Transformer decoder refines object queries $Q$, which are composed of content queries $Q_c$ and positional queries $Q_p$, through attention mechanisms to capture semantic and spatial information of instances. Typically, $Q_p$ are randomly initialised and updated along with $Q_c$. However, this is insufficient for learning the useful spatial representations of leaf instances under complex plant structures, particularly with small labelled datasets. Hence, we propose GDPQ, which dynamically generates positional queries conditioned on guide functions.

Our GDPQ is inspired by \cite{he2023dynamic}, which proposes to modulate positional queries using cross-attention maps. In \cite{he2023dynamic}, the positional queries $Q_p$ at current block $t$ are updated as:
\begin{equation}
    \label{eq:cross}
    Q_p^t = h(A^{t-1} K_p^{t-1} + B),
\end{equation}
where $h$ is an MLP, $A^{t-1}$ is the cross-attention maps (\textit{c.f.} \cref{eq:cross-attention}), $K_p^{t-1}$ denotes the positional encodings of pixel features, and $B$ is a trainable bias term.

In our case, since the trained guide functions possess instance location priors, we use them to embed the intermediate predictions of GMT and generate positional queries based on the resulting embeddings, as shown in \cref{fig:gdpq}. The proposed GDPQ is expressed as:
\begin{equation}
    \label{eq:gdpq}
    Q_p^t = h(\mathcal{E}(\mathcal{S}^{t-1}; \Psi) + B),
\end{equation}
where $\mathcal{S}^{t-1}$ denotes the set of predicted instance masks from the previous Transformer block, and $\mathcal{E}(\mathcal{S}^{t-1}; \Psi) = \bigconcat_{S \in \mathcal{S}^{t-1}} e(S; \Psi)$ represents the concatenation of all guided embeddings (\textit{c.f.} Equation (\ref{eq:ef})(\ref{eq:ge})) of these masks. The computation process of $\mathcal{E}(\mathcal{S}^{t-1}; \Psi)$ is also highlighted in \cref{fig:gdpq}. We configure $h\left(\cdot\right)$ as a three-layer MLP to map the concatenated embeddings to the same dimension as object queries (from $d_g$ to $d_p$). In this manner, the positional queries are conditioned on the guide functions, which embody positional priors, and are dynamically generated based on intermediate mask predictions, which continue to refine as the Transformer decoder progresses deeper. Finally, the positional queries generated by GDPQ are summed with the content queries to form new object queries, which are then processed by the Transformer blocks.

\mypar{Loss Function} 
The loss function of GMT is defined as:
\begin{equation}
    L = L_{\text{guide}} + L_{\text{M2F}}, 
\end{equation}
where $L_{\text{guide}}=\lambda_{\text{guide}}L_{\text{1}}$ is a weighted $L_{\text{1}}$ loss applied to the guided features, as described in GEFM and \cref{fig:auxsup}. $L_{\text{M2F}}=\lambda_{\text{ce}} L_{\text{ce}} + \lambda_{\text{dice}} L_{\text{dice}} + \lambda_{\text{cls}} L_{\text{cls}}$ follows the standard Mask2Former loss \cite{cheng2022masked}, with $L_{\text{ce}}$ and $L_{\text{dice}}$ denoting binary cross-entropy loss and Dice loss \cite{milletari2016v} for mask prediction, and $L_{\text{cls}}$ as the cross-entropy loss for classification. The $\lambda$ terms denote the weights for each corresponding loss.

\section{Experimental Results}
\label{sec:results}
In this section, we discuss the experimental details and results on three public datasets, where the proposed GMT is compared with the baseline (\ie Mask2Former) and other recent methods addressing the leaf instance segmentation task. Ablation studies are also conducted to demonstrate the efficacy of different components introduced in GMT.

\subsection{Datasets}
\mypar{CVPPP LSC} The CVPPP 2017 Leaf Segmentation Challenge dataset (CVPPP LSC) \cite{minervini2016finely} is widely used to benchmark leaf instance segmentation algorithms. We conduct experiments on the most commonly used subset, A1, which consists of a training set of 128 images and a hidden test set of 33 images. The image resolution is $500\times530$ pixels. We form a validation set by randomly selecting 12 images from the training set. Results are reported on the hidden test set.

\mypar{MSU-PID} The Michigan State University Plant Imagery Database (MSU-PID) \cite{cruz2016multi} includes Arabidopsis and bean plants. We use the Arabidopsis subset containing 576 annotated images of $116\times119$ pixel resolution, which are evenly distributed in 16 different plants. We randomly select 11, 2, and 3 different plants to form the training, validation, and test sets. Repeated experiments are performed over 3 different data splits, with average results on the test sets reported.

\mypar{KOMATSUNA} The KOMATSUNA dataset \cite{uchiyama2017easy} is formed by two subsets: one consists of RGB images and the other contains RGB-D images. Our experiments are performed on the RGB subset, which consists of 900 $480\times480$-pixel images evenly captured from 5 different plants. We divide the dataset into the training, validation, and test sets by randomly selecting 3, 1, and 1 different plants, respectively. Repeated experiments are performed over 3 different data splits, with average results on the test sets reported.

\subsection{Implementation Details}
\label{sec:implementation_details}
\mypar{Model Settings} By default, both the baseline model (Mask2Former) and our GMT use a ResNet-50 \cite{he2016deep} backbone. We evaluate other backbone options (ResNet-101 and Swin Transformer \cite{liu2021swin}) on GMT at \cref{subsec:abl}. As the number of training images of a leaf dataset is typically small, the baseline model and GMT\footnote {For GMT, we load the COCO pre-trained weights from Mask2Former, omitting the incompatiable keys.} are pre-trained on COCO instance segmentation dataset \cite{lin2014microsoft}.

\mypar{Data Augmentation} We apply same data augmentation techniques to all datasets: random horizontal and vertical flips, followed by random scaled cropping. The input image sizes for CVPPP LSC, KOMATSUNA, and MSU-PID are set to $512\times512$, $480\times480$, and $256\times256$, respectively.

\mypar{Training Strategies} When training the guide functions, we set the number of guide functions $d_{g}=16$ and the distance to separate different instances $\epsilon=2$. Let $i\in\left[0,d_{g} - 1\right]$ be the index of a guide function, the learnable parameters $\psi$ are randomly initialised as:
\begin{small}
\begin{align}
    \psi_i[1] &\sim U(0,50), \, \psi_i[2]=0, \, \psi_i[3] \sim U(0,2\pi); \, i<\frac{d_g}{2}, \notag \\
    \psi_i[1] &=0, \, \psi_i[2]\sim U(0,50), \, \psi_i[3]\sim U(0,2\pi); \, i\geq\frac{d_g}{2},
\end{align}
\end{small}
\noindent where $U(\cdot)$ denotes the uniform distribution. We use an AdamW optimiser \cite{loshchilov2017decoupled} with a 0.01 learning rate, conducting 1,000 epochs of training and selecting the functions with the minimum training error as final.

For GMT training, we set the loss weights as follows: $\lambda_{\text{ce}}=2$, $\lambda_{\text{dice}}=5$, $\lambda_{\text{cls}}=2$, and $\lambda_{\text{guide}}=5$. The number of classes of $L_{\text{cls}}$ is set to 1, to distinguish between leaves and background. We use AdamW with an initial learning rate of 0.0001 and a batch size of 12 across all datasets. Specifically, for CVPPP LSC, the training lasts 1,000 epochs with learning rate reductions by a factor of 0.1 at epochs 900 and 950. For KOMATSUNA nd MSU-PID, GMT is trained for 200 epochs, with learning rate multiplying by 0.1 at epoch 50 and 150. Models demonstrating optimal performance on the validation set are chosen as the final test models.

\mypar{Evaluation Metrics} We evaluate our approach using three standard metrics widely adopted for leaf instance segmentation \cite{scharr2016leaf}: Best Dice (BD), Symmetric Best Dice (SBD), and Difference in Count ($|$DiC$|$).

The generalised form of BD is defined as:
\begin{equation}
    \label{eq:bd}
    \operatorname{BD}(Y^a,Y^b) = \frac{1}{M} \sum_{i=1}^M \max_{1\leq j \leq N} \operatorname{DICE}(Y^a_i,Y^b_j),
\end{equation}

\noindent where $Y^a$ and $Y^b$ are two sets of instance masks, with $M$ and $N$ denoting the number of masks in $Y^a$ and $Y^b$, respectively. The \textsc{dice} score is defined as:
\begin{equation}
    \operatorname{DICE}(Y^a_i,Y^b_j) = \frac{2|Y^a_i \cap Y^b_j|}{|Y^a_i| + |Y^b_j|},
\end{equation}
\noindent where $|\cdot|$ indicates the number of pixels.

The BD between predicted instance masks $\hat{Y}$ and ground-truth instance masks $Y$ is calculated as $\operatorname{BD}(\hat{Y},Y)$, and the SBD is defined as:
\begin{equation}
    \operatorname{SBD}(\hat{Y},Y) = \min \left\lbrace \operatorname{BD}(\hat{Y},Y), \operatorname{BD}(Y,\hat{Y})\right\rbrace.
\end{equation}

The final metric, $|$DiC$|$, quantifying the difference between the predicted and ground-truth number of instances (\ie leaves), is defined as:
\begin{equation}
    |\operatorname{DiC}| = |M-N|,
\end{equation}
\noindent where $M$ and $N$ are the number of predicted and ground-truth instances, respectively; $|\cdot|$ indicates the absolute value.

BD and SBD measure instance segmentation quality. BD computes the average Dice score of the best match between predicted and ground-truth masks, but only in one direction, which may overestimate accuracy in cases of under-segmentation. For example, if the model predicts one leaf that perfectly matches a ground-truth leaf, but the ground truth contains multiple leaves, BD could still score 100. SBD offers a more balanced evaluation by considering both matching directions. Additionally, $|$DiC$|$ quantifies over or under-segmentation by comparing the predicted and ground-truth instance counts. Together, BD, SBD, and $|$DiC$|$ offer a comprehensive assessment of segmentation quality and leaf-counting accuracy.

\subsection{Quantitative and Qualitative Results}
\begin{table}[t]
  \centering
  \caption{Test set results on CVPPP LSC \cite{minervini2016finely}, MSU-PID \cite{cruz2016multi} and KOMATSUNA \cite{uchiyama2017easy}. We implement Mask2Former \cite{cheng2022masked} and our GMT. Other results for CVPPP LSC were sourced from \cite{chen2023pctrans} (they did not report BD), and those for MSU-PID and KOMATSUNA were taken from \cite{bhagat2022eff} (they did not report SBD and $|$DiC$|$).}
    \begin{adjustbox}{width=0.8\linewidth}
    \begin{tabular}{lccc}
    \toprule
          & BD $\uparrow$   & SBD $\uparrow$  & $|$DiC$|$ $\downarrow$ \\
    \midrule
    \rowcolor{gray!10} \multicolumn{4}{l}{\textit{CVPPP LSC} \cite{minervini2016finely}} \\
    ~BISSG \cite{liu2022biological} & - & 87.3  & 1.40 \\
    ~OGIS \cite{yi2021object}  & - & 87.5  & 1.10 \\
    ~PCTrans \cite{chen2023pctrans} & - & 88.7  & 0.70 \\
    ~Mask2Former \cite{cheng2022masked}  & 91.0 & 89.5  & 0.67 \\
    ~GMT (Ours)  & \textbf{91.8} & \textbf{90.1} & \textbf{0.48} \\
    \rowcolor{gray!10} \multicolumn{4}{l}{\textit{MSU-PID} \cite{cruz2016multi}} \\
    ~Yin \etal \cite{yin2017joint} & 65.2  & - & -{} \\
    ~Eff-Unet++ \cite{bhagat2022eff} & 71.2  & -{} & -{} \\
    ~Mask2Former \cite{cheng2022masked} & 85.7 & 83.1 & \textbf{0.47}\\
    ~GMT (Ours)  & \textbf{86.1} & \textbf{83.5} & \textbf{0.47} \\
    \rowcolor{gray!10} \multicolumn{4}{l}{\textit{KOMATSUNA} \cite{uchiyama2017easy}} \\
    ~Ward \etal \cite{ward2018deep} & 62.4  & -{} & -{} \\
    ~UPGen \cite{ward2020scalable} & 77.8  & -{} & -{} \\
    ~Eff-Unet++ \cite{bhagat2022eff} & 83.4  & -{} & -{} \\
    ~Mask2Former \cite{cheng2022masked} & 93.5 & 90.9 & \textbf{0.20}\\
    ~GMT (Ours)  & \textbf{94.0} & \textbf{91.0} & 0.22 \\
    \bottomrule
    \end{tabular}%
    \end{adjustbox}
  \label{tab:main_results}%
\end{table}%

The overall quantitative results on CVPPP LSC, MSU-PID and KOMATSUNA are presented on \cref{tab:main_results}, with qualitative results shown in \cref{fig:visual_results}. Additional visual results are shown in Supplementary Materials. For all three datasets, we observe that Mask2Former (COCO pre-trained) already surpass previous SOTA approaches, so we mainly compare Mask2Former with the proposed GMT.


\begin{table*}[t]
  \centering
  \begin{minipage}{0.7\textwidth}
    \caption{MSU-PID \cite{cruz2016multi} and KOMATSUNA \cite{uchiyama2017easy} test set results across different leaf sizes.}
    \label{tab:size_results}
    \begin{adjustbox}{width=\textwidth}
        \begin{tabular}{lccc|ccc|ccc}
            \toprule
            & \multicolumn{3}{c|}{Small} & \multicolumn{3}{c|}{Medium} & \multicolumn{3}{c}{Large} \\
            \midrule
            & BD $\uparrow$ & SBD $\uparrow$ & $|$DiC$|$ $\downarrow$ & BD $\uparrow$ & SBD $\uparrow$ & $|$DiC$|$ $\downarrow$ & BD $\uparrow$ & SBD $\uparrow$ & $|$DiC$|$ $\downarrow$ \\
            \midrule
            \rowcolor{gray!10} \multicolumn{10}{l}{\textit{MSU-PID} \cite{cruz2016multi}} \\
            ~Mask2Former \cite{cheng2022masked} & 73.3 & 68.3 & 0.70 & 83.6 & 78.3 & 0.51 & \textbf{58.4} & 41.9 & 0.85 \\
            ~GMT (Ours)  & \textbf{76.2} & \textbf{69.9} & \textbf{0.69} & \textbf{84.1} & \textbf{79.3} & \textbf{0.46} & 55.7 & \textbf{45.6} & \textbf{0.69} \\
            \rowcolor{gray!10} \multicolumn{10}{l}{\textit{KOMATSUNA} \cite{uchiyama2017easy}} \\
            ~Mask2Former \cite{cheng2022masked} & 80.8 & 74.5 & \textbf{0.33} & 90.0 & 88.7 & 0.14 & 97.9 & \textbf{97.6} & \textbf{0.01} \\
            ~GMT (Ours)  & \textbf{81.7} & \textbf{75.5} & \textbf{0.33} & \textbf{91.4} & \textbf{90.0} & \textbf{0.13} & \textbf{98.0} & 97.5 & 0.02 \\
            \bottomrule
        \end{tabular}%
    \end{adjustbox}
  \end{minipage}%
  \hfill
  \begin{minipage}{0.28\textwidth}
    \centering
    \caption{Performance of our method with different backbones.}
    \label{tab:varying}%
    \begin{adjustbox}{width=\textwidth}
    \begin{tabular}{lccc}
        \toprule
        Backbone    & BD $\uparrow$   & SBD $\uparrow$  & $|$DiC$|$ $\downarrow$ \\
        \midrule
        SwinT-B & \textbf{92.2} & 89.9  & 0.70 \\
        Res-101 & 92.1  & 89.3  & 0.73 \\
        Res-50 & 91.8  & \textbf{90.1} & \textbf{0.48} \\
        \bottomrule
    \end{tabular}%
    \end{adjustbox}
  \end{minipage}
\end{table*}

\begin{figure}[t]
     \centering
    \includegraphics[width=0.95\linewidth]{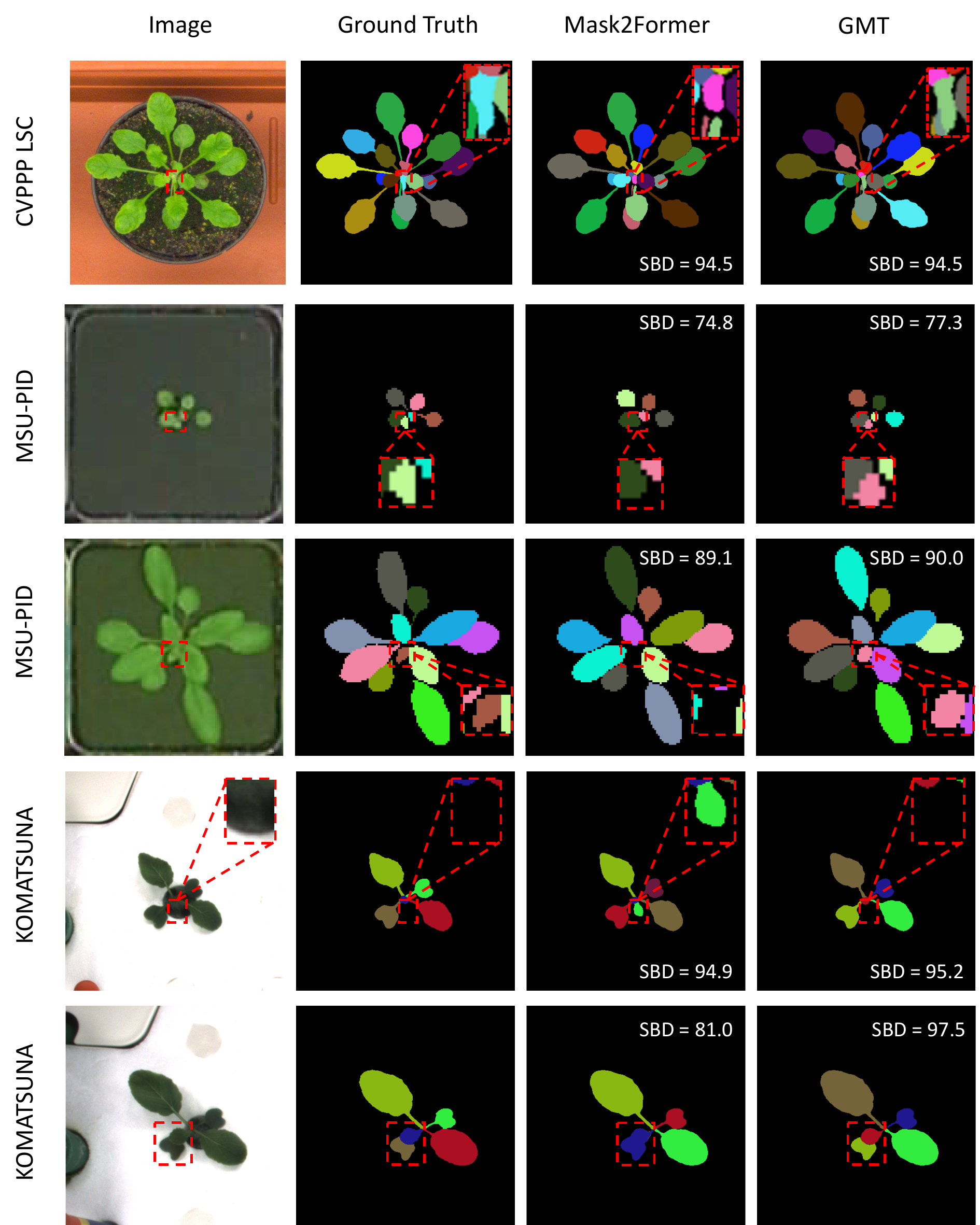}
    \caption{Qualitative results of CVPPP LSC validation set, MSU-PID and KOMATSUNA test sets. The SBD (as the major segmentation metric) of each model prediction is also displayed.}
    \label{fig:visual_results}
\end{figure}

\mypar{CVPPP LSC}
We observe consistent performance improvements of GMT compared to Mask2Former, as shown in the first part of \cref{tab:main_results}, with increases of +0.8 in BD, +0.6 in SBD and +0.19 in $|$DiC$|$. \cref{fig:visual_results} row 1 demonstrates that GMT is able to delineate the shape of small leaves at the plant centre more accurately.

\mypar{MSU-PID} \cref{tab:main_results} second part shows that GMT surpass Mask2Former by +0.4 on BD and SBD, respectively, while performing on par in $|$DiC$|$. \cref{fig:visual_results} row 2 shows that GMT mitigates under-segmentation when one tiny leaf overlaps another leaf, whereas Mask2Former merges the two leaves and misses part of the outer shape; row 3 illustrates that GMT reduces the occurrence of missing leaves.

\mypar{KOMATSUNA} As shown in the third section of \cref{tab:main_results}, GMT improves over Mask2Former with a +0.5 increase in BD and slightly outperforms it in SBD by +0.1, while showing a minor underperformance in $|$DiC$|$ by -0.02. \cref{fig:visual_results} row 4 demonstrates that GMT avoids confusing similar green objects with leaves, and row 5, again, shows that GMT reduces under-segmentation in occlusion scenarios.

\mypar{Leaves with Different Areas} To better understand which aspects of leaf instance segmentation our method improves, we classify leaves as small, medium and large based on their visible pixel areas. We evaluate Mask2Former and our GMT on the MSU-PID and KOMATSUNA test sets across these different leaf area categories. Based on the raw image resolution and the distribution of leaf areas in each dataset, the categories of leaf pixel areas are defined as:
\begin{small}
\begin{itemize}
    \item \textbf{MSU-PID:} $\text{Small}\in(0, 12^2]$, $\text{Medium}\in(12^2, 24^2]$, $\text{Large}\in(24^2, \infty)$; 
    \item \textbf{KOMATSUNA:} $\text{Small}\in(0, 35^2]$, $\text{Medium}\in(35^2, 56^2]$, $\text{Large}\in(56^2, \infty)$.
\end{itemize}
\end{small}
It is important to note that the visible area may not be equal to the actual size of a leaf (\eg when a leaf is partially covered by another), but our evaluation still provides meaningful insights into the specific improvements made by GMT.

The results shown in \cref{tab:size_results} demonstrate that for small and medium leaves on both datasets, GMT improve the segmentation quality (\ie BD and SBD) notably while its counting ability remains on par with Mask2Former. For large leaves in MSU-PID, GMT performs worse in BD but outperforms in SBD and $|$DiC$|$. As discussed in \cref{sec:implementation_details}, BD may underestimate the errors caused by under-segmentation or missing leaves due to its one-directional matching nature, while SBD is a more balanced metric for segmentation quality. In fact, the better $|$DiC$|$ score of GMT suggests that, despite a lower BD compared to Mask2Former, GMT is more effective at identifying all leaves. Finally, for large leaves in KOMATSUNA, GMT performs comparably to Mask2Former. Overall, GMT excels in segmenting small and medium leaves, which are more challenging in leaf instance segmentation, while maintaining solid performance on large leaves.

For plant phenotyping, accurate segmentation of smaller leaves is crucial as they reflect key processes in early growth stages such as photosynthesis and nutrient uptake \cite{liu2020leaf}. Smaller leaves are also more sensitive to environmental changes such as plant stress \cite{kozlov2022leaf} or water loss \cite{wang2019smaller}. Moreover, as younger and smaller leaves continually emerge, segmenting them alongside older and larger ones provides a more comprehensive understanding of plant growth dynamics \cite{vanhaeren2015journey}. Failing to capture these smaller and overlapped leaves can result in misinterpretation of plant development.

\mypar{Summary} In this section, we show that our GMT outperforms all the previous SOTA models and the baseline Mask2Former across all three datasets in overall leaf instance segmentation. Through the visual results and the evaluation on various leaf areas, we show that GMT effectively addresses the most challenging aspects of plant image analysis, including accurate segmentation of smaller, overlapped and easily missing leaves, and reducing confusion between leaves and other similar objects.

\subsection{Ablation Study}
\label{subsec:abl}
We perform two sets of ablative experiments for GMT on CVPPP LSC with results reported on the hidden test set.

Firstly, we assess different backbones with results shown in \cref{tab:varying}. We find that the smallest model (ResNet-50) achieves the best \textsc{SBD} and $|$DiC$|$, while the largest model (SwinT-B) reaches the best \textsc{BD}. As discussed earlier at \cref{sec:implementation_details}, better \textsc{SBD} and $|$DiC$|$ but worse \textsc{BD} indicate that ResNet-50 more accurately identifies and segments all leaves, while SwinT-B can produce higher-quality masks for some instances. Additionally, due to the small size of plant datasets, larger backbones tend to overfit, which is why ResNet-50 is used for all other experiments.

Next, we evaluate each component of GMT (\textit{c.f.} \cref{subsubsec:gmt}), to understand their contributions to the overall model performance. The results are shown in \cref{tab:remove}. Compared to the baseline (Mask2Former), we observe that using a single component or a combination of two tends to enhance specific aspects of the model's abilities, but not all. For instance, using GPE alone (row (a)) slightly improves BD while maintaining SBD but worsens $|$DiC$|$. However, when all three modules, GPE, GEFM, and GDPQ, are integrated, GMT outperforms Mask2Former across all metrics. This comprehensive improvement highlights the effectiveness of our overall design of combining all modules to address the complexities of leaf instance segmentation.
\begin{table}[t]
\centering
\caption{Performance of using different combinations of GPE, GEFM, and GDPQ. Mask2Former (M2F) \cite{cheng2022masked} does not incorporate any of the proposed modules.}
\label{tab:remove}
\begin{adjustbox}{width=1\linewidth}
    \begin{tabular}{l|ccc|ccc}
        \toprule
         & \textbf{GPE} & \textbf{GEFM} & \textbf{GDPQ} & \textbf{BD ↑} & \textbf{SBD ↑} & \textbf{$|$DiC$|$ ↓} \\ \hline
(a)                 & \checkmark   &        &        & 91.3          & 89.5          & 0.79          \\
(b)                &      & \checkmark   &        & 91.7          & 89.3          & 0.76          \\
(c)                &        &        & \checkmark   & 90.4          & 89.0          & 0.70          \\
(d)          & \checkmark   & \checkmark   &        & 91.5          & 89.3          & 0.73          \\
(e)          & \checkmark   &       & \checkmark   & 90.6          & 89.5          & 0.55          \\
(f)         &        & \checkmark   & \checkmark   & 91.0          & 88.0          & 0.76          \\ \hline
\textbf{GMT}        & \checkmark           & \checkmark            & \checkmark            & \textbf{91.8} & \textbf{90.1} & \textbf{0.48} \\
\textbf{M2F} \cite{cheng2022masked} &            &             &             & 91.0          & 89.5          & 0.67          \\
        \bottomrule
    \end{tabular}
\end{adjustbox}
\end{table}
\section{Conclusion}
Given the complexity of plant structures and the typically small size of labelled datasets, leaf instance segmentation in plant images is challenging. To address this task, we present GMT, a Transformer-based model that effectively incorporates leaf position priors to better capture plant complexity on limited data. We demonstrate the effectiveness of GMT in three widely used datasets, showing that GMT outperforms the SOTA (\cref{tab:main_results}), particularly in challenging cases such as smaller and overlapping leaves (\cref{tab:size_results} and \cref{fig:visual_results}). Ablation studies further validate the backbone choice and overall design of GMT. Future work will focus on extending GMT to address more complex agricultural scenarios while reducing its computational demands.

\section*{Acknowledgements}
This project was funded by the BBSRC grant BB/Y512333/1 ``PhenomUK-RI: The UK Plant and Crop Phenotyping Infrastructure'', and Microsoft Accelerating Foundation Models Research (AFMR) grant: Agricultural Foundation Models via Domain-Specific Pre-Training.

{\small
\bibliographystyle{ieee_fullname}
\bibliography{egbib}
}

\clearpage

\appendix
\section{Supplementary Material}
\subsection{Additional Visual Results}
\begin{figure}[t]
     \centering
    \includegraphics[width=\linewidth]{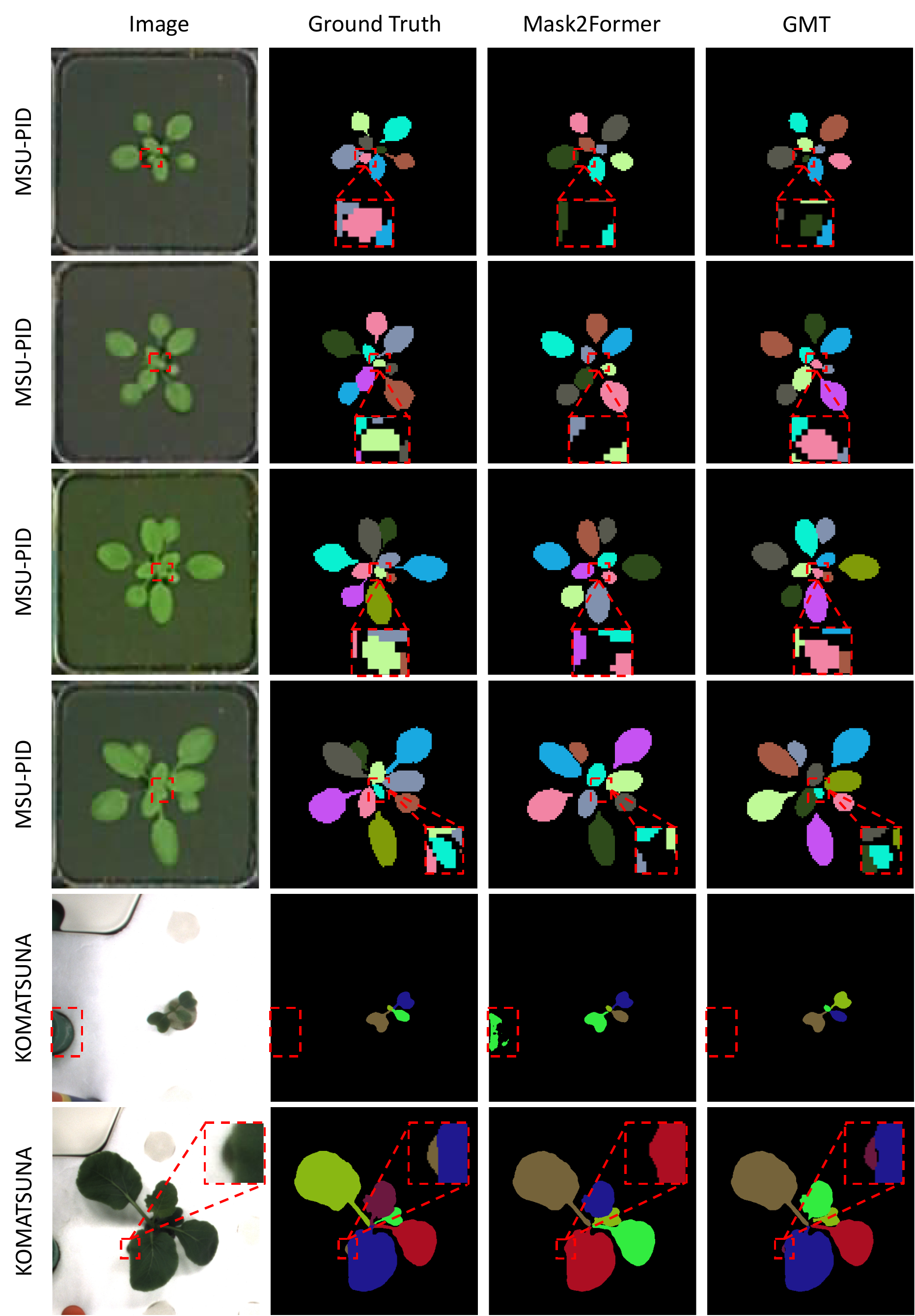}
    \caption{Additional visual results on the MSU-PID \cite{cruz2016multi} and KOMATSUNA \cite{uchiyama2017easy} test sets. Key differences between the predictions of Mask2Former \cite{cheng2022masked} and our method (GMT) is highlighted in \textcolor{red}{red boxes}.}
    \label{fig:supp_results}
\end{figure}
We provide additional qualitative results and discussion to complement Section 4.3 and Figure 5 in the main paper. 

\Cref{fig:supp_results} presents further qualitative comparisons on the MSU-PID \cite{cruz2016multi} and KOMATSUNA \cite{uchiyama2017easy} test sets, highlighting the differences between the predictions of Mask2Former \cite{cheng2022masked} and our GMT model. In \Cref{fig:supp_results}, rows 1 to 4 show that Mask2Former misses some leaves, while GMT successfully captures and segments them. Row 5 presents that GMT correctly distinguishes leaves from other green objects. Row 6 highlights GMT’s capability in segmenting overlapping leaves, even when they are small.

These findings are consistent as those in the main paper, demonstrating that GMT is able to handle challenging scenarios in plant image analysis, ultimately benefiting plant phenotyping and a variety of agricultural applications.

\end{document}